\documentclass[accepted]{article}

\usepackage{microtype}
\usepackage{graphicx}
\usepackage{subfigure}
\usepackage{booktabs} 
\usepackage{amssymb}
\usepackage{amsmath}  
\usepackage{algorithm2e}
\usepackage{tikz}
\SetInd{0.25em}{0.75em} 
\newcommand{\circled}[1]{\tikz[baseline=(char.base)]{\node[shape=circle,draw,fill=black,inner sep=1pt] (char) {\textcolor{white}{#1}};}}

\newcommand{\vvspace}{\vspace{-0.7em}}
\DeclareMathOperator{\Var}{Var}
\usepackage{hyperref}

\newcommand{\SysName}{\textup{ReSpec}\xspace}

\usepackage{xspace}
\usepackage{mathtools} 
\usepackage{bm}
\usepackage{enumitem}
\usepackage{textcomp}
\usepackage{xcolor}
\usepackage{changepage}
\usepackage{framed}

\usepackage{hyperref} 

\usepackage{graphicx}
\usepackage{subcaption}
\usepackage{tikz}

\usepackage{tabularx}
\usepackage{multirow}
\usepackage{booktabs}
\usepackage{makecell}
\usepackage[figuresleft]{rotating}
\usepackage[flushleft]{threeparttable}

\usepackage{algorithmicx}
\usepackage{algorithm}
\usepackage[noend]{algpseudocode}
\usepackage{amsmath}
\usepackage{xcolor}
\usepackage{colortbl}

\algnewcommand\algorithmicinput{\textbf{Input:}}
\algnewcommand\Input{\item[\algorithmicinput]}
\algnewcommand\algorithmicoutput{\textbf{Output:}}
\algnewcommand\Output{\item[\algorithmicoutput]}
\renewcommand{\algorithmiccomment}[1]{\hfill$\vartriangleright$\textit{#1}}

\usepackage[utf8]{inputenc} 
\usepackage[newfloat=true]{minted} 
\usepackage{tcolorbox} 
\usepackage{soul} 
\usepackage{pifont} 
\usepackage{bbding} 
\usepackage[textsize=footnotesize, linecolor=magenta, bordercolor=magenta,
    backgroundcolor=white, textwidth=40pt]{todonotes}
\usepackage{marginnote}

\algnewcommand{\LeftComment}[1]{\Statex \(\triangleright\) #1}

\usepackage{mlsys2025}

\begin{document}

\twocolumn[
\mlsystitle{\SysName: Towards Optimizing Speculative Decoding in Reinforcement Learning Systems}



\mlsyssetsymbol{equal}{*}
\begin{mlsysauthorlist}
\mlsysauthor{Qiaoling Chen}{ntu,sqj}
\mlsysauthor{Zijun Liu}{thu,sqj}
\mlsysauthor{Peng Sun}{sqj}
\mlsysauthor{Shenggui Li}{ntu}
\mlsysauthor{Guoteng Wang}{sqj}
\mlsysauthor{Ziming Liu}{nus,sqj}
\end{mlsysauthorlist}

\begin{mlsysauthorlist}
\mlsysauthor{Yonggang Wen}{ntu}
\mlsysauthor{Siyuan Feng}{sii,sqj}
\mlsysauthor{Tianwei Zhang}{ntu}
\end{mlsysauthorlist}

\mlsysaffiliation{ntu}{Nanyang Technological University, Singapore}
\mlsysaffiliation{sqj}{Shanghai Qiji Zhifeng Co., Ltd., Shanghai, China}
\mlsysaffiliation{thu}{Tsinghua University, Beijing, China}
\mlsysaffiliation{nus}{National University of Singapore, Singapore}
\mlsysaffiliation{sii}{Shanghai Innovation Institute, Shanghai, China}



\mlsyskeywords{Machine Learning, MLSys}

\vskip 0.3in

\begin{abstract}
Adapting large language models (LLMs) via reinforcement learning (RL) is often bottlenecked by the generation stage, which can consume over 75\% of the training time. Speculative decoding (SD) accelerates autoregressive generation in serving systems, but its behavior under RL training remains largely unexplored. We identify \textbf{three critical gaps} that hinder the naïve integration of SD into RL systems: diminishing speedups at large batch sizes, drafter staleness under continual actor updates, and drafter-induced policy degradation. 

To address these gaps, we present \SysName, a system that adapts SD to RL through three complementary mechanisms: dynamically tuning SD configurations, evolving the drafter via knowledge distillation, and weighting updates by rollout rewards. On Qwen models (3B–14B), \SysName achieves up to $4.5\times$ speedup while preserving reward convergence and training stability, providing a practical solution for efficient RL-based LLM adaptation.
\end{abstract}

]

\printAffiliationsAndNotice{}


\section{Introduction}

\begin{figure*}[tb]
    \centering
\includegraphics[width=0.98\linewidth]{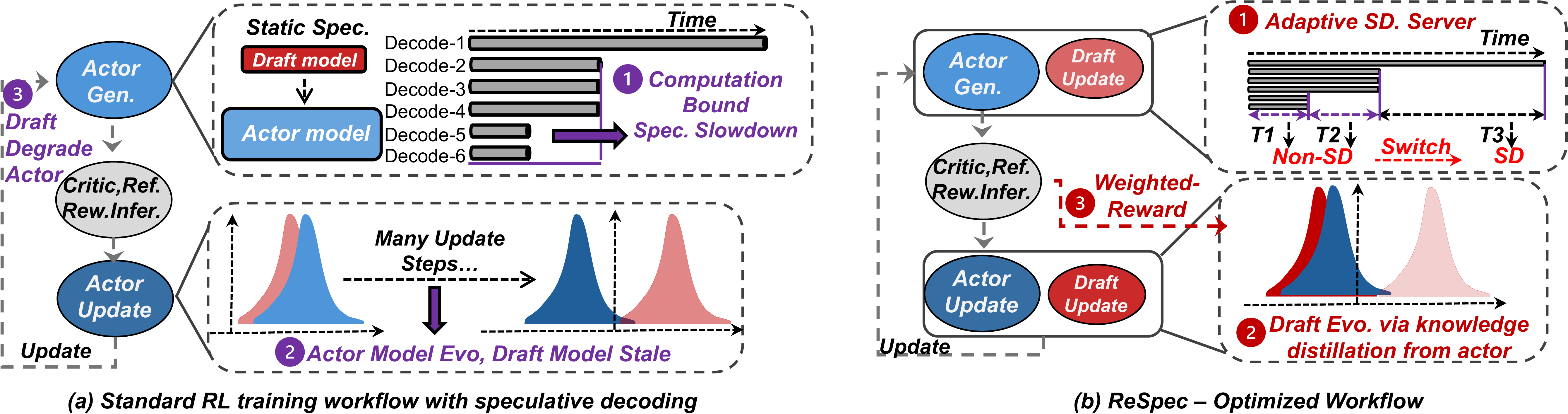}
    \caption{Overview of SD in RL training and our proposed \SysName system.
    (a) The RL training workflow integrated with SD, highlighting three fundamental \textbf{gaps}.    (b) The design of \textbf{\SysName}, which addresses these gaps through three complementary mechanisms: \circled{1} dynamically tuning SD configurations to match workload conditions, \circled{2} continuously aligning the drafter with the evolving actor using on-policy signals, and \circled{3} weighting drafter updates by rollout quality to maintain stable policy learning. Together, these mechanisms enable stable and efficient SD throughout RL training.}
    \label{fig:workflow}
    \vvspace{}
\end{figure*}

Reinforcement learning (RL) has become a practical route to adapt large language models (LLMs) for complex, high-level objectives, such as improved alignment, reasoning, and domain-specific skills \cite{qwen3,gemini,Claude,llama4}. An RL training iteration typically consists of three stages: \emph{generation}, \emph{inference}, and \emph{training}. In the generation stage, the actor model produces rollout trajectories for a batch of prompts via autoregressive decoding; the inference stage evaluates these trajectories using auxiliary models such as reward or critic networks; the training stage consumes the resulting signals to update the actor parameters. Among these stages, \textbf{generation is consistently the dominant bottleneck} \cite{StreamRL}.
As shown in Table~\ref{tab:RLtime}, for LLMs trained with a maximum response length of 8K tokens, generation accounts for up to 86\% and 75\% of the wall-clock iteration time in math and code models, respectively.

This bottleneck is further exacerbated by the algorithms widely used in practice \cite{deepseekmath,qwen2math}. Classic policy optimization methods such as PPO~\cite{ppo1,ppo2} combine trajectory-level rewards with critic-based feedback to stabilize updates, while more recent group-based approaches, e.g., GRPO~\cite{deepseekmath} and DAPO~\cite{dapo}, generate multiple candidate completions per prompt and compute updates from relative comparisons inside each group. Although group sampling improves exploration and gradient quality, it also multiplies the number of decoded tokens per prompt, further inflating the cost of the generation stage.

A natural optimization to address this bottleneck is \emph{speculative decoding} (SD) \cite{SD1,SD2,miaoSDServey}. Standard auto-regressive decoding requires one expensive forward pass of the target model per token. SD reduces this cost by introducing a lightweight \emph{drafter} that proposes short token sequences, which are then validated by the target model in parallel. If the drafter and target align closely, many drafted tokens can be accepted at once, thereby reducing the number of target forwards per generated token. SD has already been widely adopted in LLM serving systems (e.g., SGLang~\cite{sglang}, TensorRT-LLM~\cite{TensorRTSD}) and commercial platforms (e.g., OpenAI~\cite{OpenAISD}, AWS SageMaker~\cite{AWS}).  Among various SD variants, \textbf{EAGLE-3}~\cite{eagle3} represents the current state of the art, achieving the highest reported speedup among lossless SD methods~\cite{medusa,lookaehad,hydra}.
Accordingly, we adopt EAGLE-3 as the default SD algorithm in our analysis, implementation, and experiments.

\textbf{Research Gaps.}
Despite the widespread success of SD in serving systems, its behavior under RL training has never been systematically examined.
Through our analysis of RL workflows and the prototype integration of SD into the generation stage, we identify three fundamental gaps that explain why naïve application of SD fails to deliver consistent gains in RL training, as illustrated in Figure~\ref{fig:workflow}(a).
(G1) \textit{Diminishing speedups at large decoding batch sizes.}
When generation already runs at high GPU utilization with large batch sizes, the marginal parallelism provided by SD becomes limited and can even be offset by drafting and synchronization overheads~\cite{dsd}. Consequently, a single static SD configuration cannot provide reliable speedups across diverse RL workloads, as shown in Figure~\ref{fig:eagle_speedup}.
(G2) \textit{Drafter staleness under continual actor updates.}
As the actor (i.e., target) model evolves with each policy update, a fixed drafter rapidly becomes misaligned with the actor distribution, leading to the loss of acceleration benefits, as observed in Figure~\ref{fig:acceptlen}.
(G3) \textit{Drafter-induced degradation of actor performance.}
Even when token-level acceptance remains correct, multi-token drafts can have large variance. Moreover, this variance increases with drafter staleness (G2), causing a higher ratio of impoverished trajectories. In RL settings, this shift degrades downstream rewards and provides misleading gradients during policy optimization, as demonstrated in Figure~\ref{fig:rewards}.

\noindent\textbf{Opportunities.}
Beyond these gaps, we identify two key opportunities in the RL generation stage that can be leveraged to make SD both stable and efficient.
First, the \emph{generation stage exhibits strong skewness}: within a batch, most sequences terminate early while a small fraction continue much longer, resulting in a time-varying active batch size. Because the efficiency of SD depends critically on the batch size, this skewness offers an opportunity to adapt SD configurations dynamically rather than relying on static hyperparameters, as shown in Figures~\ref{fig:reqsplot} and~\ref{fig:speedup}.
Second, RL rollouts inherently produce rich, \emph{on-policy diagnostic signals} that are routinely logged but never reused, including per-step target, draft logits, and rewards. These signals provide a valuable, untapped supervision source for maintaining drafter alignment with the evolving actor policy, enabling continuous adaptation without introducing additional overhead to the sampling process.

Motivated by the above-mentioned gaps and opportunities, we introduce \textbf{\SysName}, a system that adapts speculative decoding to the unique demands of RL training. \SysName provides three targeted mechanisms, each directly addressing one of the gaps (Figure~\ref{fig:workflow} (b)). Specifically, (1) for G1, we design \textit{Adaptive Speculative Decoding Server}. It dynamically selects and applies speculative-decoding configurations based on lightweight profiling and runtime workload signals. (2) For G2, we introduce \textit{Drafter Evolution via Knowledge Distillation} with a \emph{Online Learner}. To keep the drafter aligned with the continually updated actor, we evolve the drafter using on-policy distillation from the actor. This distillation pipeline continuously transfers the actor’s distributional knowledge into the lightweight drafter so that drafted proposals remain relevant as the policy changes. (3) For G3, we design \textit{Reward-Weighted Adaptation}. To avoid inducing distributional shifts that harm policy learning, we weight drafter updates by rollout quality, which prioritizes high-reward trajectories during adaptation. This reward-weighted mechanism ensures the drafter favors trajectories that preserve or improve downstream rewards, reducing the risk of misleading gradients during RL optimization.

\noindent\textbf{Our contributions are summarized as follows:}
\begin{itemize}[leftmargin=*, itemsep=0pt, topsep=0pt, label=\ding{72}]
\item We conduct the first systematic study of SD in end-to-end RL training of LLMs. Our analysis reveals three critical gaps, including \emph{diminishing speedups}, \emph{drafter staleness}, and \emph{policy degradation}, that explain why naïve SD integration fails to provide consistent acceleration.

\item We identify two underexplored opportunities in RL generation: (1) \emph{skewed generation workloads} that enable dynamic adaptation of SD configurations, and (2) \emph{on-policy diagnostic signals} that can be repurposed as supervision for continuous drafter alignment.

\item We design \textbf{\SysName}, the first system that adapts SD to RL training. \SysName{} comprises two tightly coupled components: an \emph{Adaptive SD Server} that performs profiling-guided and runtime-tuned SD scheduling, and an \emph{Online Learner}, which continuously maintains drafter alignment through three complementary mechanisms: evolving the drafter via knowledge distillation from the actor, weighting updates by rollout rewards to preserve policy quality, and performing asynchronous, replay-buffered updates to avoid blocking generation.

\item We implement and evaluate \SysName{} on Qwen models (3B–14B), showing that it preserves training stability and reward convergence while achieving up to \textbf{$4.5\times$} speedup over standard RL training, demonstrating its practicality for RL training pipelines.

\end{itemize}
\section{Background and Related Work}
\subsection{Reinforcement Learning for LLMs}
\begin{table}[t]
\centering
\begin{tabular}{llll}
\hline
 & Generation & Inference & Training \\
\hline
Math & 83\%--86\% & 1.7\%--2.4\% & 12.3\%--14.6\% \\
Code & 70.9\%--75.5\% & 11.4\%--14\% & 13.1\%--15.1\% \\
\hline
\end{tabular}
\caption{Time breakdown during RL iterations for 7B models (max response length 8K tokens).}
\label{tab:RLtime}
\vvspace
\end{table}

\begin{figure*}[tb]
    \centering
    \begin{minipage}[t]{0.69\linewidth}
        \centering
        \includegraphics[width=\linewidth]{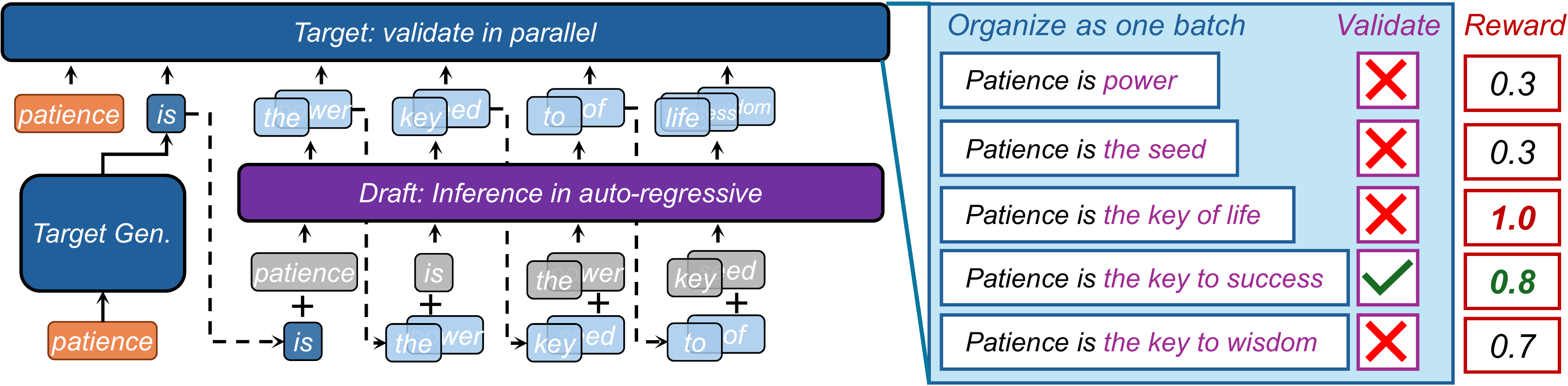}
        \caption{EAGLE-3 workflow. The target model generates one token, while the draft model produces multiple candidates using hidden states. The target verifies all candidates in a single forward pass and accepts four tokens (“the key to success”) with only two forward passes.}
        \label{fig:eagle3}
    \end{minipage}
    \hfill
    \begin{minipage}[t]{0.29\linewidth}
        \centering
        \includegraphics[width=\linewidth]{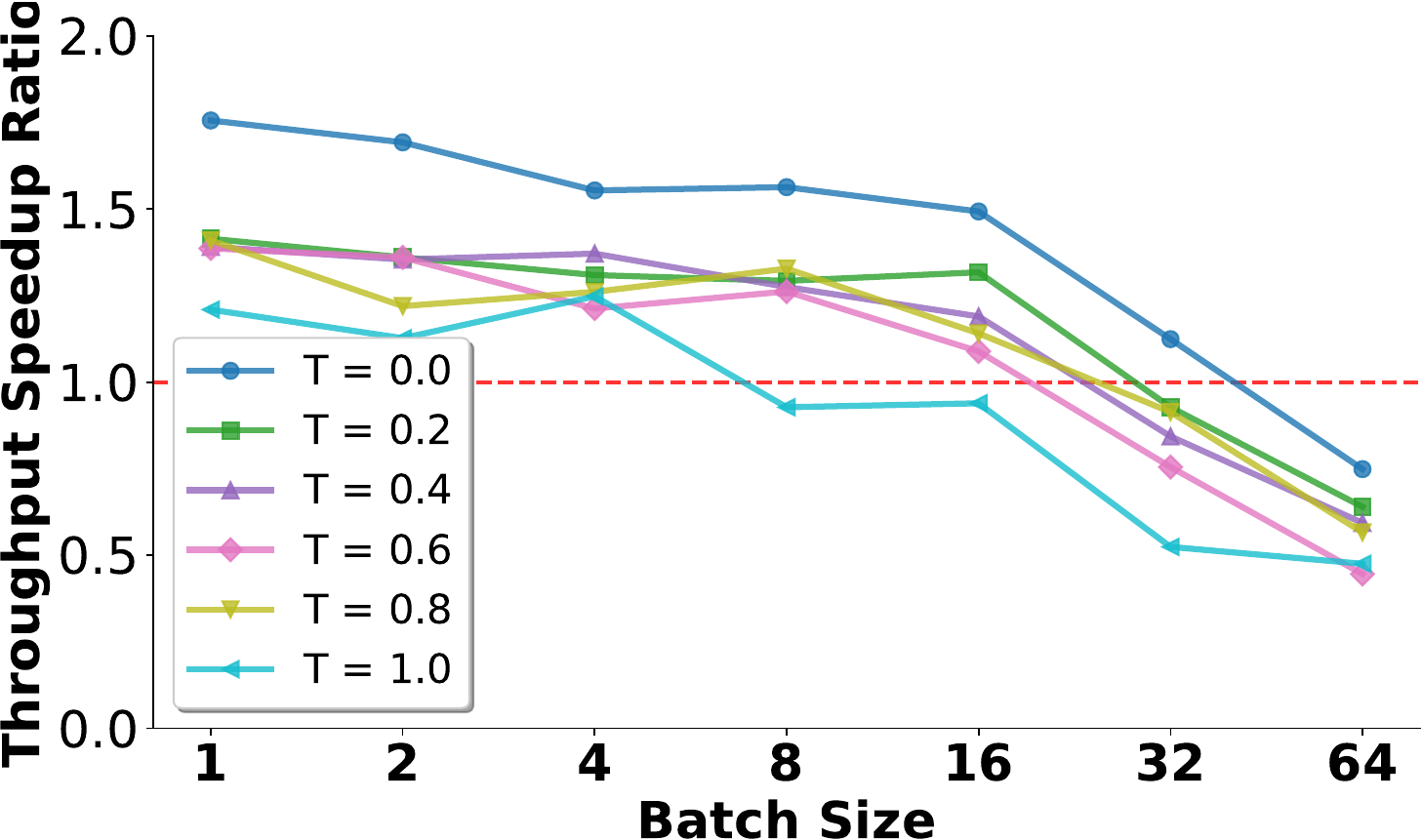}
        \caption{Speedup ratio under different batch sizes and temperature on H100 and Qwen2.5-7B-instruct for the MTBench dataset.}
        \label{fig:eagle_speedup}
    \end{minipage}
    \vvspace
\end{figure*}

\noindent\textbf{Workflow.}
RL for LLMs typically proceeds in iterations that consist of three stages: generation, inference, and training. In the \emph{generation} stage, an actor model produces rollout trajectories for a batch of prompts. This stage contains a prefill (prompt encoding) step followed by autoregressive decoding of tokens until termination. In the \emph{inference} stage, auxiliary models (reference, reward, critic) evaluate the generated rollouts by performing forward passes to produce scalar signals or logits used by the update. Finally, in the \emph{training} stage, the actor consumes these signals to compute losses and apply parameter updates. 

\noindent\textbf{Algorithms.}
Classic policy optimization methods such as PPO~\cite{ppo1,ppo2} combine trajectory-level rewards with critic-based, action-level feedback to reduce the variance of updates. Recent production-oriented recipes, e.g., GRPO~\cite{deepseekmath}, DAPO~\cite{dapo}, leverage \emph{group-based} sampling: for each prompt, the actor generates a group (many candidate completions) and updates are computed from relative comparisons inside the group. Group sampling improves the exploration and gradient signal in model optimization but increases the number of decoded tokens per prompt, amplifying the cost of the generation stage and motivating efforts to accelerate decoding without altering the sampling distribution.

\noindent\textbf{Bottleneck.}
The generation stage consistently dominates the end-to-end RL iteration time. As shown in Table~\ref{tab:RLtime}, for LLMs trained with a maximum response length of 8K tokens, generation accounts for up to 86\% and 75\% of the total time for math and code models, respectively. This imbalance highlights that decoding, rather than model updating or reward evaluation, is the primary performance bottleneck in practical RL pipelines.

\noindent\textbf{Existing Frameworks.} Numerous frameworks aim to accelerate RL post-training for LLMs. Early systems~\cite{nemo, openrlhf, puzzle, deepspeedchat} focus on orchestrating multi-stage RL workflows, while later efforts introduce architectural and scheduling optimizations—e.g., hybrid or multi-controller designs~\cite{verl, distflow}, fused or parallelized stages~\cite{rlhfuse, realhf}, and asynchronous post-training for long-tail rollouts~\cite{areal, rollpacker, rhymesrl}. However, these frameworks largely overlook the generation stage and its optimization opportunities. In contrast, \SysName explicitly targets this bottleneck by integrating SD into RL training.


\subsection{Speculative Decoding}

Standard autoregressive decoding requires one target-model forward per token, which is computationally expensive. Speculative decoding (SD) mitigates this by using a lightweight \emph{drafter} to propose token sequences, which are then verified by the expensive \emph{target} model in batches\cite{SD1,SD2}.
Subsequent works extend this paradigm along multiple dimensions: Medusa~\cite{medusa} increases proposal parallelism, Cascade Speculative Drafting~\cite{cascade} introduces multi-stage chaining, and EAGLE-series methods~\cite{eagle,eagle2,eagle3} progressively enhance feature utilization, draft-tree construction, and parallelization, achieving state-of-the-art inference efficiency.

\noindent\textbf{Speculative decoding procedure.}
We adopt a representative speculative sampling procedure similar to the recent SOTA method EAGLE-3~\cite{eagle3} as our reference implementation. 
Starting from a realized prefix \(T_{1:j}\), SD repeats a two-stage cycle. 
(1) \emph{Drafting:} the drafter autoregressively proposes a block \(\hat T_{j+1:j+k}=(\hat t_{j+1},\ldots,\hat t_{j+k})\), where $\hat{t}$ represents individual tokens; 
(2) \emph{Validation:} the target evaluates the proposed block in a batched forward and accepts drafted tokens sequentially according to an acceptance test; if a token is rejected, it is replaced by sampling from a residual distribution derived from the target logits.

\noindent\textbf{Acceptance rule (single-token).}
For a single drafted token \(\hat t\) with drafter probability \(q(\hat t)\) and target probability \(p(\hat t)\), the acceptance probability is (omitting conditions on prefixes):
\begin{equation}
\Pr\left[\text{accept }\hat t\right] \;=\; \min\!\left(1,\; \frac{p(\hat t)}{q(\hat t)}\right).
\label{eq:accept_prob}
\end{equation}
If rejected, a replacement token is sampled from the residual distribution:
\begin{equation}
r(x) = \frac{\max(0,\,p(x)-q(x))}{\sum_y \max(0,\,p(y)-q(y))}, 
\label{eq:residual}
\end{equation}
where $x,y$ denote single tokens. It guarantees that the marginal sample follows \(p\)~\citep{SD1}.




\noindent\textbf{Multi-token drafting and accumulated mismatch.}
When drafting multiple tokens sequentially, each proposed token depends on the previously accepted tokens. We calculate the variance of the acceptance probability $\frac{p}{q}$ for sampled sequences as:
\begin{equation}
    \Var_{T \sim q} \left[\prod_{t \in T} \frac{p(t)}{q(t)}\right] = \prod_{t \in T} \left(1+D_{\chi^2} (p(t)||q(t)) \right) -1,
\end{equation}
where $D_{\chi^2}$ is the Chi-squared divergence. If $D_{\chi^2} (p(t)||q(t))$ is approximated by a constant $\delta$, the variance $(1+\delta)^{|T|}-1$ increases exponentially with the sequence length. As the target model distribution evolves, this divergence typically widens, amplifying the variance of acceptance probabilities: some tokens become increasingly unlikely to be drafted ($\frac{p(t)}{q(t)}\uparrow$), while others are drafted often but rarely accepted ($\frac{p(t)}{q(t)}\downarrow$).

\noindent\textbf{Acceptance length.} 
A key practical metric in SD is the \emph{acceptance length}, 
defined as the number of consecutive tokens in a drafted block that are accepted by the target model before the first rejection. Longer acceptance lengths generally lead to higher speedups because more tokens can be validated per forward step, but also increase the likelihood of multi-token distribution mismatch, as described above.

\noindent\textbf{Speculative decoding cost model.}
Let \(C_p\) denote the average per-token cost of the \textbf{target model} under naïve decoding (i.e., one forward pass per token), and \(C_q\) denote the average per-token cost of the \textbf{drafter model}, with \(C_q \ll C_p\).
Suppose we draft blocks of \(k\) tokens, where the average acceptance rate per token is \(r\), i.e., the expected fraction of drafted tokens ultimately accepted.
If validating a block of \(k\) tokens requires target-model computation roughly proportional to \(k\), but can be parallelized with an efficiency factor \(\alpha\), then the effective target cost for validating the entire block is approximately \(kC_p / \alpha\).
Under these assumptions, the expected cost per accepted token in SD is estimated as: 
\begin{equation}
\mathrm{Cost}_{\text{SD}} \approx \frac{kC_q + kC_p/\alpha}{r k} \;=\; \frac{C_q + C_p/\alpha}{r}.
\label{eq:cost_model}
\end{equation}

\section{Challenges of Applying SD in RL}
Applying SD inside RL training is attractive because the generation stage frequently dominates the iteration time (Table~\ref{tab:RLtime}). However, prior work on online SD has focused on inference-serving scenarios, where draft models are updated online to track shifting request distributions~\cite{distillspec,OSD}. In contrast, RL training requires the drafter to adapt to the evolving actor policy itself, posing three practical gaps not addressed by existing methods.

\noindent\textbf{GAP 1: diminishing speedup at large batch sizes.} In RL training, we enlarge the batch size to fit GPU utilization in the decoding phase. However, when the batch size in decoding is already large, GPUs are typically operating near high utilization through straightforward batching of independent sequences. Therefore, the extra parallelism that SD provides yields little marginal benefit. Besides, SD introduces additional overheads, as illustrated in Equation \ref{eq:cost_model}. This added draft cost and verification synchronization offset, and even exceeded the speedup. As illustrated in Figure \ref{fig:eagle_speedup}, speedups from SD shrink as batch size increases.

\begin{figure}[tb]
    \centering
    \includegraphics[width=0.98\linewidth]{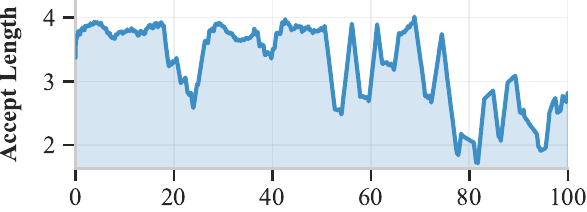}
    \caption{Acceptance length of the draft model during 100 RL steps with Qwen2.5-7B-instruct model and math dataset. 
    As training progresses, the EAGLE-3 drafter quickly becomes stale and its acceptance length drops.} 
    \label{fig:acceptlen}
    \vvspace
\end{figure}

\noindent\textbf{GAP 2: SD staleness during RL training.} 
In RL training, the actor parameters are continually updated. 
A drafter distilled from an earlier snapshot may become misaligned with the evolving actor, 
leading to lower acceptance length and thus diminishing the benefits of SD. 
Figure~\ref{fig:acceptlen} illustrates this effect: the acceptance length of the EAGLE-3 drafter decreases 
as training advances.

\begin{figure}[tb]
    \centering
    \includegraphics[width=0.98\linewidth]{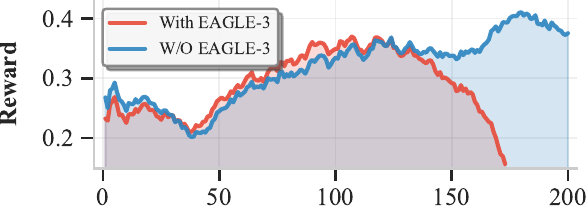}
    \caption{Evolution of rewards over 200 RL steps for Qwen2.5-7B on the math dataset. Naïve application of EAGLE-3 leads to a measurable drop in reward, illustrating drafter-induced distributional bias during RL training.}
    \label{fig:rewards}
    \vvspace
\end{figure}

\noindent\textbf{GAP 3: drafter-induced degradation of the actor's performance.}
In RL, returns are a nonlinear, trajectory-level function: small changes in an early token can cascade to very different continuations and hence markedly different rewards. Although the SD acceptance test (Eq.~\eqref{eq:accept_prob}) preserves the \emph{marginal} token distribution at the single step, it remains high variance in acceptance probability over multi-token continuations induced by the target model and its parameter updates. As a result, multi-token drafts that are accepted sequentially can be impoverished and systematically shift the distribution of completed rollouts away from what the actor would have produced under naive decoding.\vvspace

This shift matters for learning. When accepted drafts contain phrases or patterns that yield lower downstream reward than the actor’s own optimal continuations, those drafter-originated sequences appear in the buffer of rollouts used for updates and thus alter the computed returns and policy gradients. Over successive updates, the actor can therefore receive biased training signals that reinforce the drafter’s suboptimal tendencies, leading to a measurable degradation of the actor’s performance. This effect is distinct from pure variance amplification: it introduces a \emph{bias} in the training distribution that can systematically pull the policy toward worse behavior.

Figure~\ref{fig:eagle3} illustrates a concrete instance: a drafter may repeatedly propose a superficially plausible phrase that yields lower downstream reward than the actor's optimal continuation (e.g., “the key to success” vs. “the key of life”), and these accepted drafts accumulate in the replay buffer. Empirically, we observe that naively applying EAGLE-3 in our RL runs can produce a measurable drop in reward after on the order of 100 update steps (Figure~\ref{fig:rewards}), consistent with drafter-induced distributional bias harming learning.

\section{Key Insights}
\label{sec:motivation}

\begin{figure}[tb]
    \centering
    \includegraphics[width=0.98\linewidth]{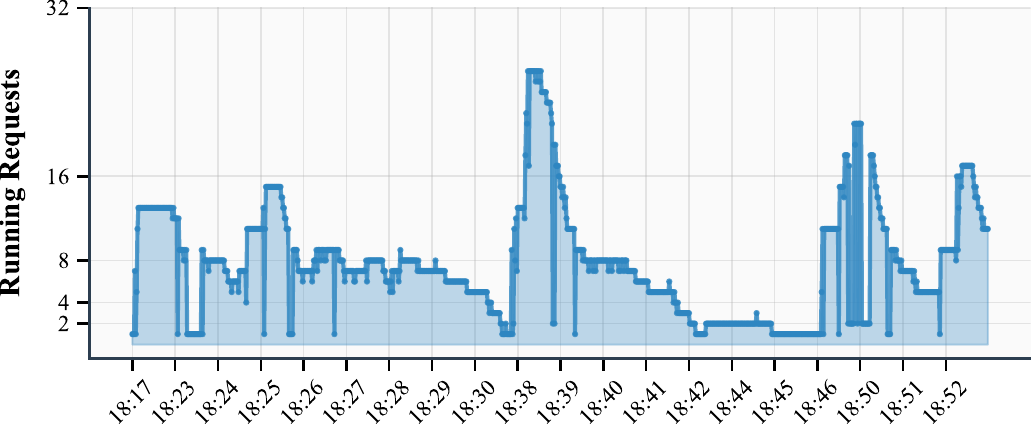}
    \caption{Number of active sequences 
    during decoding in Qwen2.5-7B RL training (Math). It illustrates the skewed generation workload: most sequences finish quickly while a few persist, causing fluctuations in effective batch size.}
    \label{fig:reqsplot}
    \vvspace
\end{figure}

\begin{figure}[tb]
    \centering
    \includegraphics[width=0.98\linewidth]{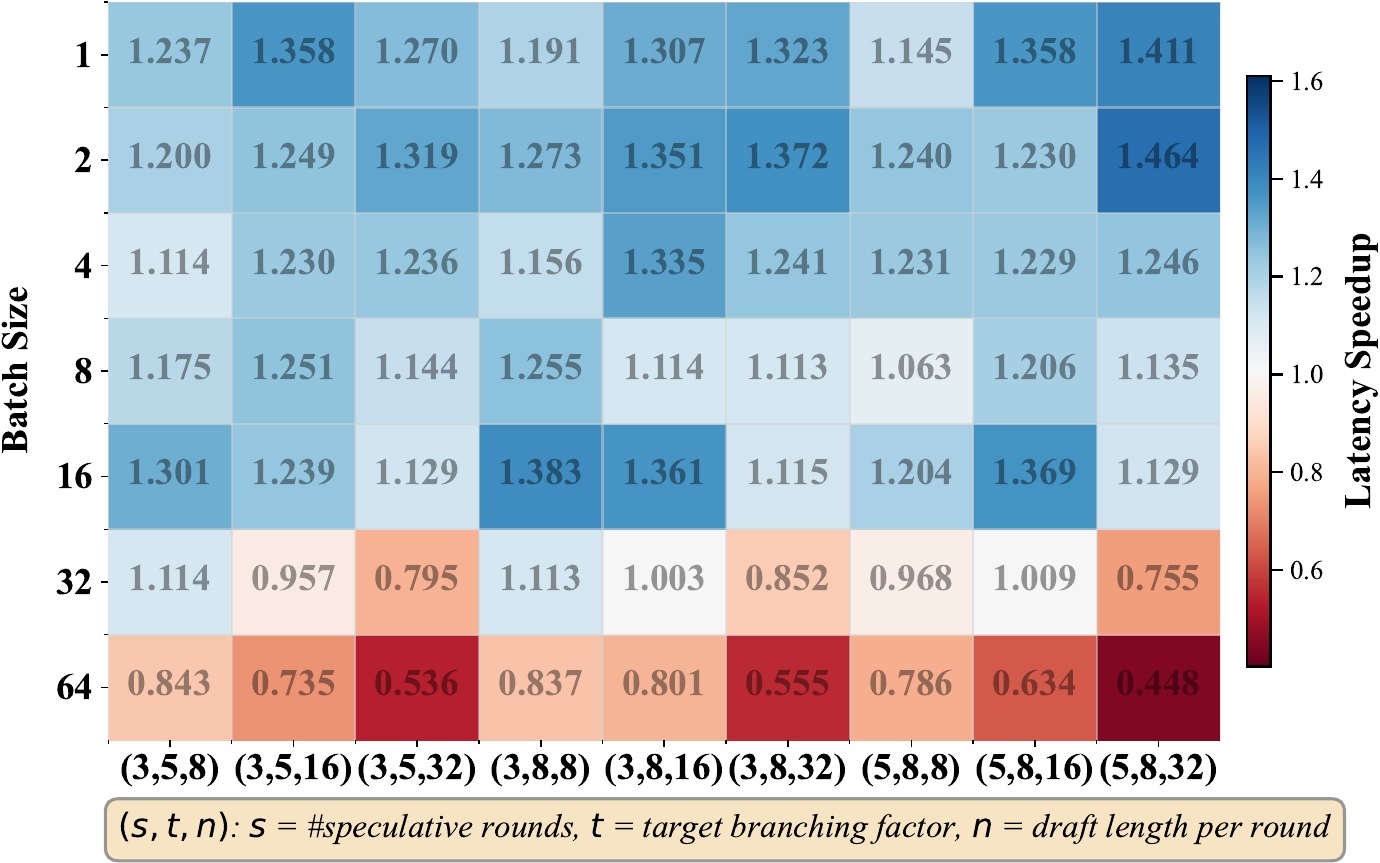}
    \caption{Latency speedup of EAGLE-3 (temperature = 0.6) on Qwen2.5-7B-instruct for the MTBench dataset. Each cell shows speedup for a specific $(s,t,n)$ configuration, showing that optimal SD depends strongly on active batch size.}
    \label{fig:speedup}
    \vvspace
\end{figure}

\subsection{Generation-stage Skewness and Its Consequences for Speculative Decoding}

We observe that the generation stage in RL training exhibits pronounced \emph{skewness}: within a single decoding batch, most sequences finish quickly while a small fraction continue for much longer. Figure~\ref{fig:reqsplot} shows this pattern concretely for Qwen-2.5–7B: the number of active requests fluctuates heavily, occasionally spiking above 16, spending much of the run near 8, and then decaying to a long tail near 1. Because autoregressive decoding is sequential, these short-lived sequences drop out early, and the active batch size decreases as decoding proceeds, producing a highly non-uniform workload and intermittent under-utilization of compute.

Crucially, we find that this time-varying active batch size substantially changes the cost–benefit trade-offs of SD. SD performance is governed by three key hyperparameters: (1) the number of speculative rounds $s$,  
(2) the target branching factor $t$, i.e., number of candidate branches validated in parallel, and  
(3) the draft length per round $n$.  Importantly, our measurements reveal that the optimal SD configuration is not fixed, but depends strongly on the \emph{active batch size}. When the batch size is small, aggressive drafting (large $s,n,t$) can yield clear gains, whereas at larger batch sizes the same configuration may even hurt performance due to overheads outweighing the benefits. For instance, in our experiments (shown in Figure \ref{fig:speedup}), the best SD configuration at batch size $2$ achieves $1.46\times$ speedup, while the same setting degrades to $0.76\times$ at batch size $32$. 

This coupling highlights that static SD hyperparameters cannot deliver consistent benefits across the dynamic regime of RL training. Instead, adaptive strategies that \emph{tune SD parameters online based on the current active batch size} are required to fully realize the potential acceleration.

\subsection{On-policy Signals and Knowledge Distillation for Drafter Alignment}

Knowledge distillation provides a principled way to align the predictive distribution of a smaller \emph{student} model with that of a larger \emph{teacher}. Prior studies \cite{distillspec,OSD} have shown that this framework is particularly effective for SD in online serving, where incoming request distributions shift over time and the draft model must continually adapt. In these settings, distillation is applied online to fit the varying data distribution.  

In contrast, SD within RL training presents a fundamentally different challenge. The request distribution is relatively stable, but the \emph{target model itself} evolves during training. This dynamic teacher–student relationship means that static draft models quickly become misaligned, leading to reduced acceptance rates and diminished decoding efficiency.  

Fortunately, we observe an underexploited opportunity in the verification step of SD: during validation, the system has access to \emph{dense, on-policy diagnostic signals}, such as the target model's per-step logits, drafter's contemporaneous log-probabilities, and scalar trajectory rewards. These signals are produced naturally during generation and reflect the actor's current behavior under the actual sampling policy, which enables the construction of distillation objectives that directly align the draft model with the current target model, effectively tracking its evolving distribution. By exploiting such information, we can continuously refine the draft model to increase accept length and stabilize RL training.

\section{System Design}

\subsection{System Overview}

\begin{figure}[tb]
    \centering
    \includegraphics[width=0.98\linewidth]{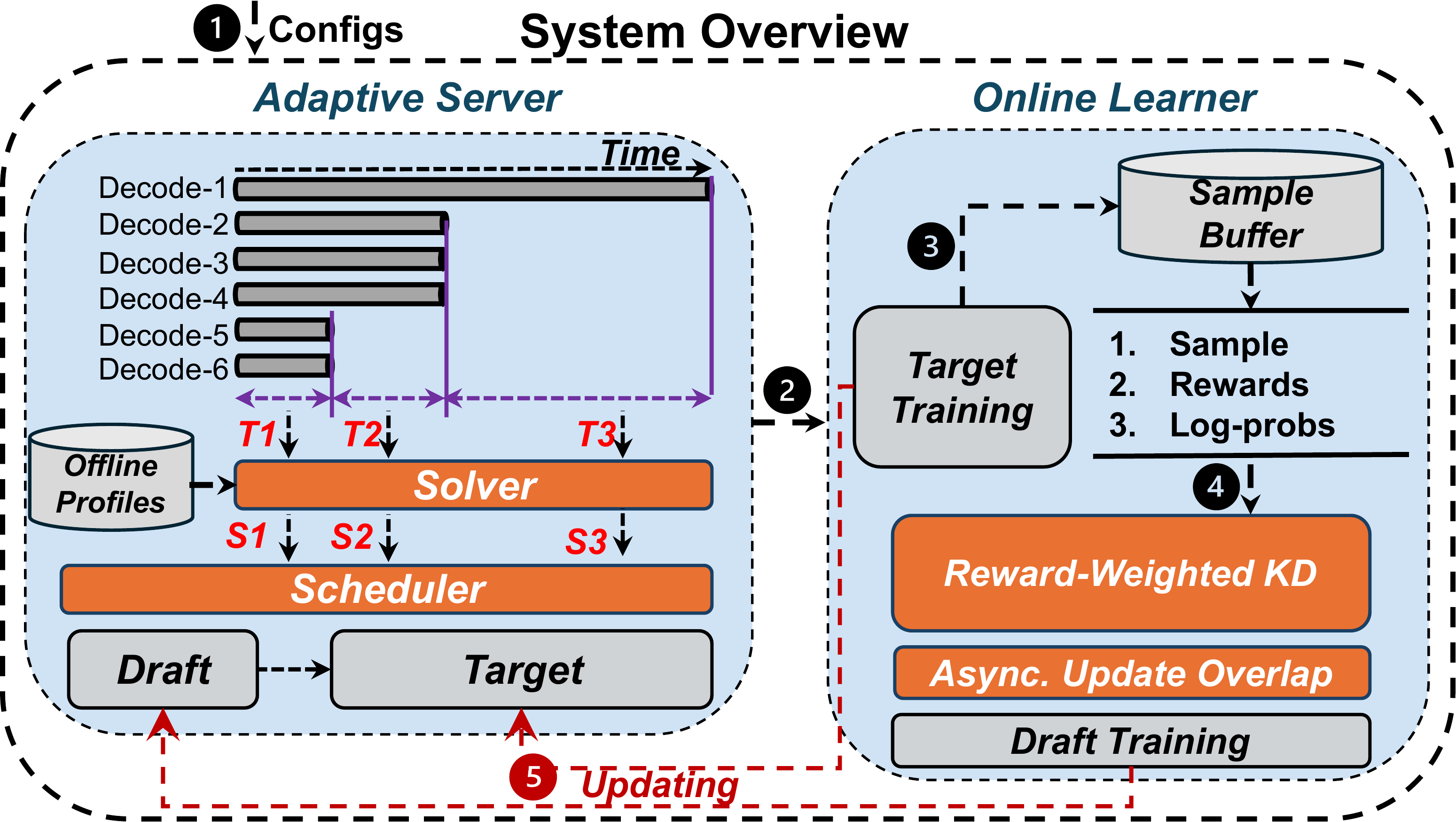}
    \caption{Workflow of \SysName during RL training. The system iteratively generates responses, stores rollout samples with associated rewards, updates the drafter via reward-weighted knowledge distillation with asynchronous overlap, and applies improved speculative decoding in the next RL step to accelerate and stabilize training.}
    \label{fig:overview}
    \vvspace
\end{figure}

Inspired by the above insights, we present \SysName, a novel system that efficiently adapts SD to RL. Figure~\ref{fig:overview} presents its architecture, which consists of two major components:

(1) The \textbf{Adaptive Server} accelerates the generation stage with SD. It integrates two modules:  
t he \emph{Solver}, which searches for and selects the optimal SD hyperparameter combinations, and the \emph{Scheduler}, which monitors the active batch size at runtime and dynamically switches SD configurations while managing the KV cache.  

(2) The \textbf{Online Learner} maintains the draft model through online training. It incorporates three modules: The \emph{Reward-Weighted Knowledge Distillation algorithm}, which distills knowledge from the evolving target model to keep the draft model aligned and ensure high acceptance length across training steps, and the \emph{Async Update Overlap} mechanism, which overlaps draft model training with RL generation.

\noindent\textbf{Workflow.}  
\circled{1} \SysName starts with a user-specified configuration, such as batch size and temperature. The Solver provides optimal SD parameters $(s,t,n)$ for the given decoding conditions, and the Scheduler adjusts them online without additional overhead.  
\circled{2} The target model generates responses and updates its weights following the standard RL training loop.  
\circled{3} Generated samples, along with log-probabilities and rewards, are stored in the Sample Buffer.  
\circled{4} The Online Learner selectively draws samples to train the draft model using the reward-weighted KD algorithm, with Async Update Overlap strategies to maximize efficiency.  
\circled{5} Updated draft weights are pushed back to the inference engine, enabling the next RL training step to proceed with improved speculative decoding.

\begin{figure}[tb]
    \centering
    \includegraphics[width=0.98\linewidth]{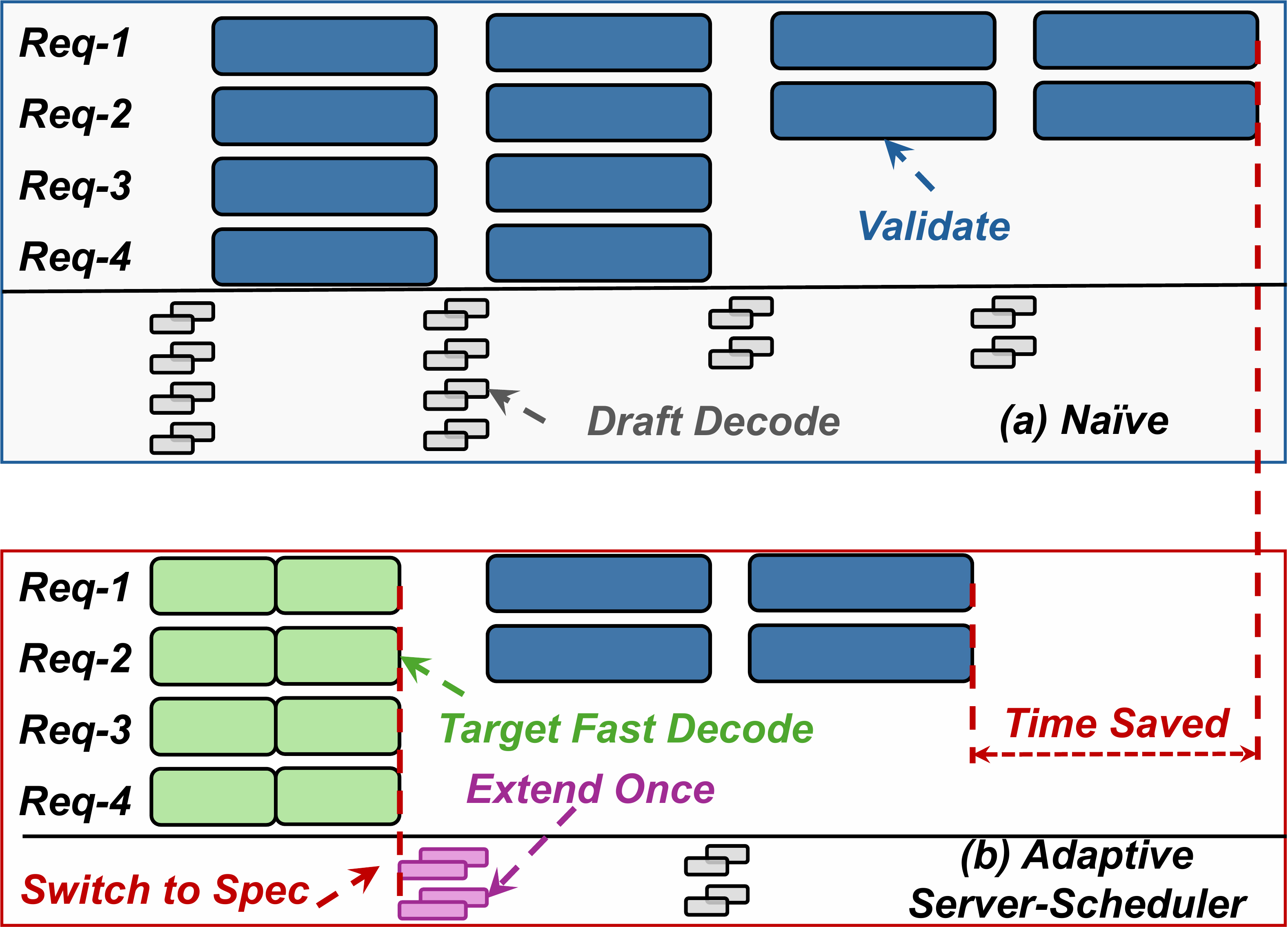}
\caption{
Workflow comparison between standard speculative decoding (a) and the proposed Adaptive Server (b).
}
    \label{fig:adaptive_server}
    \vvspace
\end{figure}

\subsection{Adaptive Server}

SD accelerates generation by drafting multiple candidate tokens in parallel and verifying them against the target model. 
However, its efficiency depends heavily on runtime conditions such as the active batch size, which fluctuates significantly during RL training. 
When GPU utilization is already saturated (e.g., large active batch sizes), enabling SD can add redundant overhead; conversely, when batch sizes shrink, SD offers substantial speedup. 
To balance these dynamics, \SysName{} introduces an \emph{Adaptive Server} that dynamically switches between speculative and non-speculative execution modes according to the current system state.

Figure~\ref{fig:adaptive_server}(a) shows the standard SD workflow without adaptation: for a batch size of 4, each request independently undergoes draft and validate stages, resulting in redundant compute when the GPU is already well utilized. 
In contrast, Figure~\ref{fig:adaptive_server}(b) depicts the adaptive mode: when the batch size is large, the Scheduler opts for pure target decoding to maintain throughput; 
as the active batch size drops (e.g., from 4 to 2), the system transitions to the speculative state. 
During this transition, the Scheduler extends the previous non-speculative decoding state to build an updated KV cache, then resumes normal speculative decoding to maximize parallelism at lower utilization. 

The Adaptive Server consists of two key components: 
(1) a \textbf{Solver} that searches for efficient SD configurations via offline profiling, 
and (2) a \textbf{Scheduler} that controls runtime switching between modes with minimal overhead.

\paragraph{Solver: Offline Profiling Guided Search.}
To determine efficient SD configurations, \SysName{} employs a profiling-based solver. 
As shown in Figure \ref{fig:speedup}, the runtime performance of SD is highly 
sensitive to the choice of hyperparameters, including the number of speculative rounds $s$, branching factor 
$t$, and number of drafted tokens per round $n$. The optimal configuration varies across different batch 
sizes due to the evolving active batch during RL training.

We first conduct an offline profiling stage, where we benchmark the execution time of draft and target models under 
various SD hyperparameter settings and batch sizes. The results are used to fit a predictive performance model that 
estimates the throughput speedup of different SD configurations as a function of the active batch size. This profiling is 
lightweight and performed once before RL training.
At runtime, the solver consults the offline-derived performance model and the observed active batch size to 
dynamically select the SD configuration expected to maximize speedup.

\paragraph{Scheduler.} The Scheduler is the runtime controller of the Adaptive Server. It determines when and how SD should be activated during generation, while minimizing system overhead. This is non-trivial in RL training, where active batch sizes shrink rapidly due to data skewness, and the benefits of SD vary across decoding stages. Switching between speculative and non-speculative execution incurs overhead. To achieve zero-overhead runtime switching, we model the decoding phase as a two-state process: \emph{spec-enabled} and \emph{non-spec}. Each request carries a lightweight flag indicating whether SD is active. The Scheduler tracks the state of the current batch and applies lightweight transitions:
\begin{itemize}[leftmargin=*, itemsep=0pt, topsep=0pt]
    \item \textbf{Non-spec $\to$ Spec.} When the Solver advises switching on SD, the Scheduler promotes the current decode batch into a spec-enabled batch by reusing the \emph{prefill} interface. This allows the draft model to generate candidates without modifying the decoding kernel.
    \item \textbf{Spec $\to$ Non-spec.} If SD is no longer beneficial, the Scheduler simply discards speculative metadata (e.g., cached candidates) and continues with regular decoding.
\end{itemize}

The Scheduler operates in a closed loop with the Solver. At runtime, it monitors active batch size and uses these signals to decide whether SD should remain enabled. 

\subsection{Online Learner}
Algorithm~\ref{alg:rl_spec} summarizes the online learner workflow used throughout our experiments. During rollouts, we run SD with the current draft $(q_\theta)$ and verify each candidate with the target model $(p)$. For every rollout, we store the input and response together with the target logits and the scalar reward. The online learner extracts distillation targets from the rollout buffer, accumulates them in a replay buffer $(Q)$, and performs periodic reward-weighted updates every $(I)$ iterations. These updates (i) keep the draft aligned with the target distribution to preserve high acceptance rates in SD, and (ii) bias the draft toward behaviors that empirically yield higher RL reward.

\begin{algorithm}[t]
\small
\caption{RL-Integrated Speculative Draft Update}
\label{alg:rl_spec}
\KwIn{Target model $(p(\cdot\mid x))$, draft model $(q_\theta(\cdot\mid x))$,
rollout buffer $(\mathcal{B})$ storing tuples ((x, y, $\log p$, r)),
update interval (I), replay buffer (Q).}

\textbf{Initialization:} Pre-train $(q_\theta)$ with offline KD on warmup data; set $Q \gets [], i \gets 0$;

\While{RL training not converged}{
Sample a mini-batch $((x, y, \log p, r))$ from rollout buffer $(\mathcal{B})$\;

Run speculative decoding with draft $(q_\theta)$ and verify outputs using target (p)\;

Collect token-level alignment info $((\log q_\theta, \log p))$ for each position\;

Store $((x, y, \log q_\theta, \log p, r))$ in replay buffer (Q)\;

$i \gets i+1$\;

\If{\(i \bmod I = 0\)}{
    \(\mathcal{L}_{\mathrm{tot}} \gets 0\)\;
    \For{each sample \((x, y, \log q_\theta, \log p, r) \in Q\)}{
        Construct soft targets \(p(\cdot\mid x,y_{<t}) \leftarrow \operatorname{softmax}(\log p)\)\;
        
        Compute per-sample reward-weighted KD loss:\;
        
      \scriptsize\(
      \mathcal{L}_{\mathrm{KD}}(x,y) = 
      w(r)\sum_{t=1}^{T}
      \mathrm{KL}\!\Big(
        \tilde p(\cdot\mid x,y_{<t})
        \;\Big\|\;
        q_\theta(\cdot\mid x,y_{<t})
      \Big)\)\;
        
        \(\mathcal{L}_{\mathrm{tot}} \gets \mathcal{L}_{\mathrm{tot}} + \mathcal{L}_{\mathrm{KD}}(x,y)\)\;
    }
    Update draft parameters with accumulated loss:
    
    \(
      \quad \theta \;\gets\; \theta - \eta\,\nabla_\theta \mathcal{L}_{\mathrm{tot}}
    \)
    
    Reset replay buffer: \(Q \gets []\)\;
}

}
\vvspace
\end{algorithm}

\subsubsection{Reward-Weighted Knowledge Distillation}
A central mismatch between RL and SD is that the target $(p)$ evolves, whereas classical KD assumes a fixed teacher. Treating all collected rollouts equally (standard KD) can therefore pull the draft toward low-quality behavior because many rollouts are erroneous or low-reward.

We perform \textbf{reward-weighted KD}. For a rollout sample $((x,y))$ with per-step target distributions $({p(\cdot\mid x,y_{<t})}$ and scalar reward $(r)$, we minimize the sample-wise loss:
\[
\mathcal{L}_{\mathrm{KD}}(x,y) \;=\; w(r)\sum_{t=1}^{T}\mathrm{KL}\!\big(p(\cdot\mid x,y_{<t}) \;|\; q_\theta(\cdot\mid x,y_{<t})\big).
\]
where $(w(r)>0)$ is a reward-based weighting function (by default $(w(r)=r)$; in practice we normalize and clip $(w(r))$ for stability). Using stored logits $(\log p)$, we materialize soft targets via $(\operatorname{softmax}(\log p))$ at update time and compute the weighted cross-entropy against the current draft $(q_\theta)$. Note that gradients are always computed with respect to the current draft $(q_\theta)$, not the recorded $(\log q_\theta)$.

\begin{figure}[tb]
\centering
\includegraphics[width=0.98\linewidth]{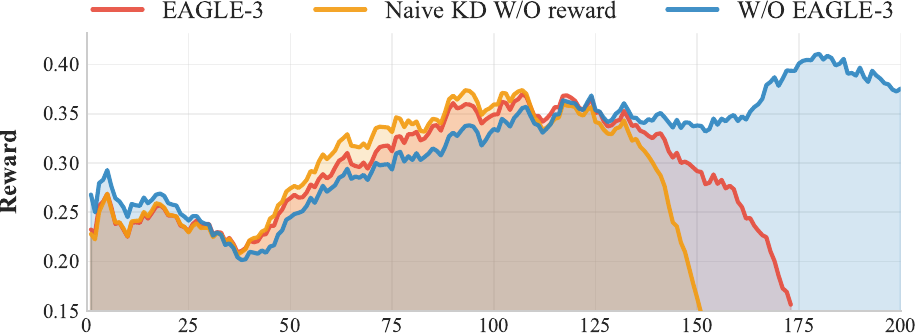}
\caption{Qwen2.5-7B reward trajectories under three strategies: No-reward KD (standard KL), Eagle-only (no draft updates), and Reward-weighted KD (ours). Reward-weighted KD prevents the rapid collapse seen in baselines.}
\label{fig:kd_compare}
\vvspace
\end{figure}

We compare three variants: (i) \textbf{No-reward KD}: standard KL on all stored samples with no weighting; (ii) \textbf{Eagle-only}: SD applied, but the draft is \emph{never} updated; (iii) \textbf{Reward-weighted KD}. On Qwen2.5-7B both no-reward KD and eagle-only begin to degrade around step $(\approx125)$. The no-reward KD run collapses fastest (near-zero reward by $(\approx150))$, while eagle-only degrades more slowly (near-zero by $(\approx175))$. Reward-weighted KD maintains steadily increasing rewards over the same horizon (see Figure~\ref{fig:kd_compare}).

\noindent\textbf{Mechanism.}
Naive KD treats all rollouts equally, so noisy or low-reward trajectories pull the draft away from the target's high-quality modes. Because SD uses the draft to accelerate rollouts, a biased draft increases the probability of generating further low-quality rollouts, creating a \emph{positive feedback loop} that degrades policy quality. Reward-weighted KD mitigates this by attenuating the contribution of low-reward samples and amplifying high-reward ones, steering the draft toward behavior that matches the target and empirically produces high reward.

\subsubsection{Async Update Overlap}
\begin{figure}[tb]
\centering
\includegraphics[width=0.98\linewidth]{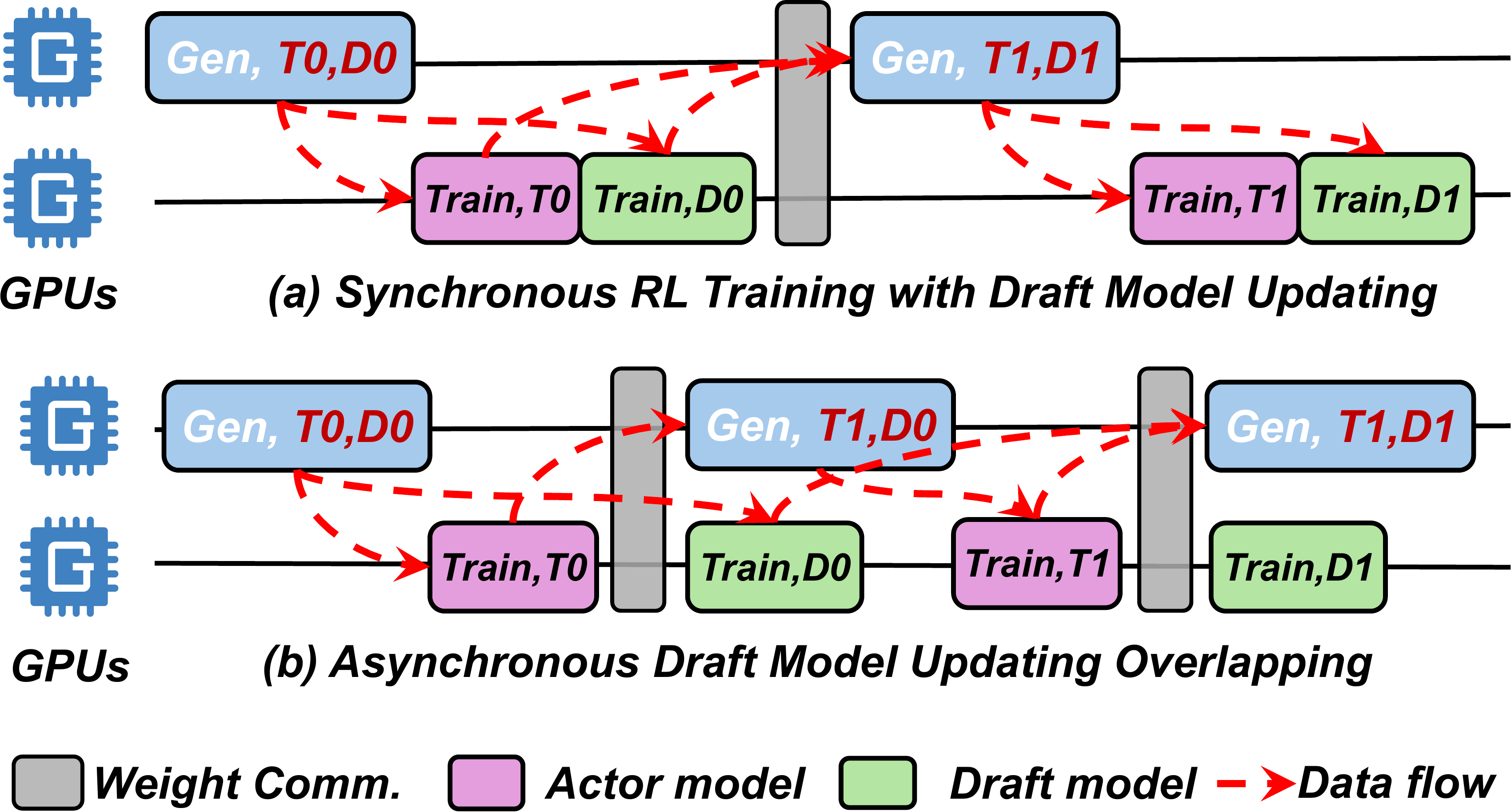}
\caption{Asynchronous draft model updating overlapping: draft updates are scheduled to run during pipeline idle slots to avoid extra wall-clock overhead.}
\label{fig:asyncupdateoverlap}
\vvspace
\end{figure}

Updating a draft model during RL training introduces unique system inefficiencies. In a naive design, draft updates are executed synchronously after each generation step, which creates pipeline bubbles: generation must wait for optimization to finish, leading to under-utilized hardware, as illustrated in Figure \ref{fig:asyncupdateoverlap} (a). Moreover, as the draft is smaller than the target, training it at the same frequency and scale as the target model is both computationally heavy and algorithmically unnecessary, given the scaling-law mismatch.

To address these challenges, we design an \emph{asynchronous update overlap} mechanism. The core idea is twofold:
(1) We maintain a replay buffer that accumulates distillation targets over multiple rollouts. The draft model is then updated every $I$ iterations using only $1/I$ of the buffer data, which amortizes optimization cost and reduces the burden of frequent updates while keeping the draft approximately aligned with the target.
(2) We overlap draft model updates with the generation stage by exploiting idle slots in the RL pipeline. Specifically, draft training runs in parallel with target rollouts during these bubbles, ensuring that updates incur negligible additional wall-clock latency, as shown in Figure \ref{fig:asyncupdateoverlap} (b).

This design balances system efficiency and algorithmic freshness: by avoiding blocking synchronization while still providing frequent enough updates, the draft model can continuously track the target’s evolving distribution without slowing down overall training. 

\section{Evaluation}
To systematically evaluate \SysName, we structure our experiments around the following research questions:
\begin{itemize}
    \vspace{-0.65em}
    \item RQ1: Can \SysName maintain stable training and reward convergence while applying SD in RL? (\S~\ref{subsec:end_to_end_evaluation})
    \item RQ2: How much end-to-end speedup does \SysName provide, and how does it scale with model size? (\S~\ref{subsec:end_to_end_evaluation})
    \item RQ3: What are the contributions of each component of \SysName to training acceleration and stability? (\S~\ref{subsec:speedup_breakdown})
    \item RQ4: How does asynchronous update frequency affect drafter alignment and policy performance? (\S~\ref{subsec:asynchronization_analysis})
\end{itemize}

\begin{figure*}[tb]
    \centering
    \includegraphics[width=0.98\linewidth]{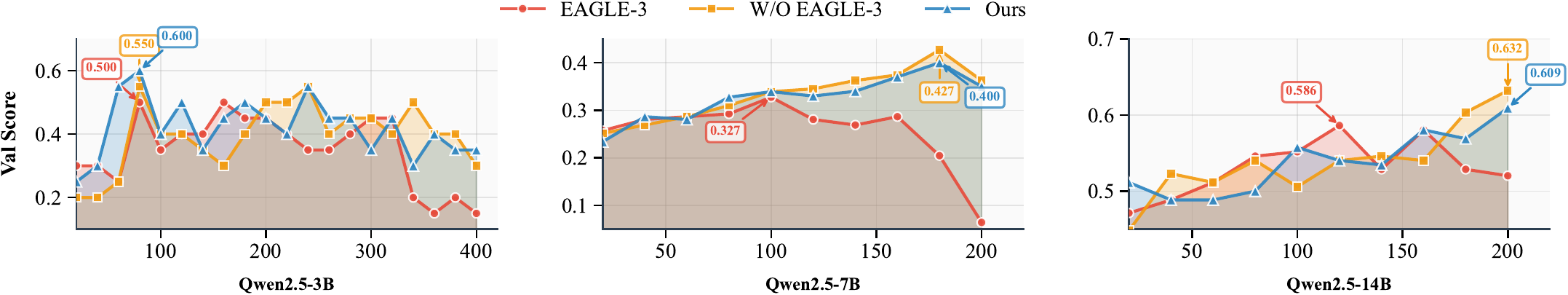}
    \caption{Validation score comparison across Qwen-2.5 models (3B–14B). 
    \SysName{} closely tracks the baseline without acceleration, while direct EAGLE-3 integration destabilizes training and leads to significant accuracy degradation.}
    \label{fig:exp_val}
    \vvspace
\end{figure*}

\begin{figure*}[tb]
    \centering
    \includegraphics[width=0.98\linewidth]{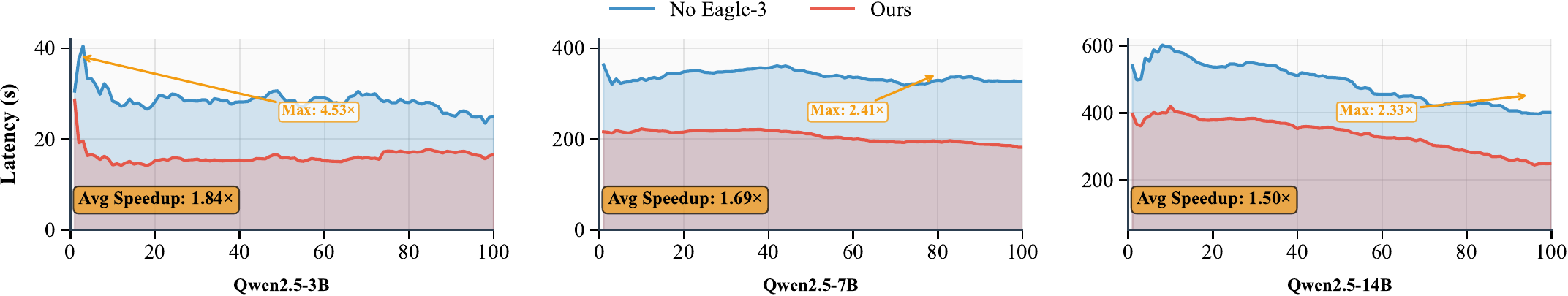}
    \caption{End-to-end training latency speedup. 
    The curves are smoothed using a moving average with a window of 20 to reduce noise. 
    \SysName{} achieves consistent $1.5\times$–$4.5\times$ acceleration across models while preserving convergence stability, unlike EAGLE-3 which fails to sustain training.}
    \label{fig:exp_speed}
    \vvspace
\end{figure*}

\begin{table}[tb]
\centering
\setlength{\tabcolsep}{4pt} 
\begin{tabular}{lcccc}
\hline
Model & Gen GPU & Train GPU & Temp. & Seq Len \\ \hline
Qwen2.5-3B   & 1       & 1         & 0.6           & 3072    \\
Qwen2.5-7B   & 8       & 8         & 0.6           & 8192    \\
Qwen2.5-14B  & 8       & 8         & 0.4           & 8192    \\ \hline
\end{tabular}
\caption{Configurations for various models.}
\label{tab:exp_config}
\vvspace
\end{table}

\noindent\textbf{Implementation.}
We build \SysName{} on top of VeRL \cite{verl} and SGLang \cite{sglang}. The implementation consists of roughly 2K lines of code (LOC), with about 500 LOC devoted to the Adaptive Server and 1500 LOC for the Online Learner. To maximize the efficiency of draft model training, we adopt tailored parallelization strategies between the actor and draft models. In addition, we implement an optimized EALGE-3 training backend to support training draft models for diverse target architectures.

\noindent\textbf{Testbed.}
All experiments are conducted on two compute nodes, each equipped with 8× NVIDIA H100 GPUs (80GB) interconnected with 900 GB/s NVLink, and connected across nodes via 8× 400 Gbps RoCE. The software stack includes PyTorch 2.7.1, Python 3.12, and CUDA 12.6.

\noindent\textbf{Models and Algorithms.}
We conduct a thorough evaluation on Qwen-2.5 models ranging from 3B to 14B parameters, trained on real-world math datasets. Detailed configurations are summarized in Table~\ref{tab:exp_config}. Training is based on the GRPO algorithm \cite{deepseekmath}, a widely adopted RL approach for LLM post-training that underpins systems such as DeepSeek-R1 \cite{dpskR1}.

\noindent\textbf{Metrics.}
We measure training efficiency via end-to-end wall-clock latency. To assess training performance, we report validation scores every 20 steps and track the mean reward throughout the RL updates.

\subsection{End-to-end Evaluation}
\label{subsec:end_to_end_evaluation}
We evaluate our method in an end-to-end RL setting, focusing on two key aspects: (i) training stability, measured by validation score during the RL updates, and (ii) efficiency, measured by wall-clock speedup relative to the baseline training without acceleration.

In terms of accuracy, we observe that \SysName{}’s Online Learner maintains stable validation scores comparable to the baseline without speculative acceleration, as shown in Figure \ref{fig:exp_val}. Across models of different scales, our method avoids the collapse when directly applying EAGLE-3 in RL training. For instance, in Qwen-3B, our validation score trajectory remains aligned with the no-acceleration baseline even after 400 steps, while EAGLE-3 often causes degradation, e.g., dropping to 0.15 after step 400. A similar pattern is found in Qwen-7B, where our method sustains higher scores (0.4 at steps 160-180) compared to EAGLE-3’s early collapse (0.06–0.2). For larger models like Qwen-14B, \SysName{} consistently tracks the baseline trends, whereas EAGLE-3 exhibits divergence after longer horizons, with unstable or declining validation scores. These results confirm that our method preserves convergence guarantees while still benefiting from SD.

In terms of efficiency, \SysName{} achieves substantial wall-clock acceleration as illustrated in Figure \ref{fig:exp_speed}. For Qwen-3B, our approach yields up to $4.53\times$ end-to-end training speedup, with a median improvement of around $1.84\times$, particularly in the early generation stages where speculative execution is most effective. For larger models, the gains remain substantial yet more stable: $1.69\times$ on average for Qwen-7B (peaking at $2.41\times$), 
and $1.50\times$ for Qwen-14B (up to $2.60\times$). 
We also observe that speculative acceleration is most pronounced during the early generation stages, 
where token-level parallelism is maximally exploitable. 

Taken together, these findings highlight that \SysName{} delivers the desired balance: it matches baseline RL training in accuracy and reward stability while offering consistent end-to-end speedups across different model sizes. This makes our method suitable for practical RL training pipelines, where both stability and throughput are critical.

\subsection{Speedup Breakdown}
\label{subsec:speedup_breakdown}

\begin{figure}[tb]
    \centering
    \includegraphics[width=0.9\linewidth]{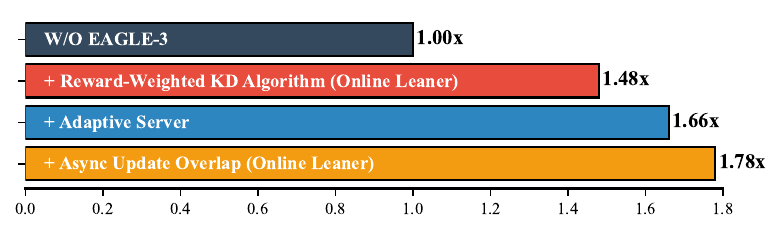}
    \caption{
    End-to-end training speedup breakdown of \SysName{} on Qwen-14B.
    Starting from the baseline without SD.
    }
    \label{fig:speedup_breakdown}
    \vvspace
\end{figure}

We further decompose the end-to-end training acceleration of \SysName{} on the Qwen-14B model to quantify the contribution of each component. 
As shown in Figure~\ref{fig:speedup_breakdown}, the baseline system without SD is normalized to 1.0$\times$. 
Introducing the \textit{Reward-Weighted KD} algorithm in the Online Learner achieves a 1.48$\times$ improvement by stabilizing SD and enhancing draft acceptance.
Adding the \textit{Adaptive Server}, including the solver and scheduler, brings an additional 12\% gain (1.66$\times$ total) by dynamically selecting the most efficient SD configuration under varying workloads.
Finally, enabling the \textit{Async Update Overlap} mechanism further hides draft update latency by overlapping online learning with rollout execution, reaching an overall 1.78$\times$ acceleration.
These results demonstrate that the algorithmic and system-level optimizations in \SysName{} are complementary and jointly contribute to end-to-end efficiency.

\subsection{Asynchronization Analysis}
\label{subsec:asynchronization_analysis}
\begin{figure}[tb]
    \centering
    \includegraphics[width=0.98\linewidth]{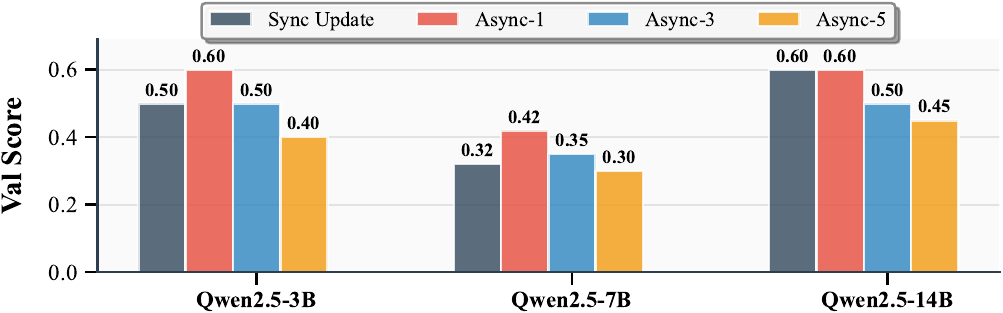}
    \caption{Validation score comparison under synchronous and asynchronous updates. Async-1 yields the best performance, while longer intervals (Async-3, Async-5) cause reward degradation, especially in smaller models.}
    \label{fig:exp_async}
    \vvspace
\end{figure}

We next evaluate the effectiveness of our asynchronous update overlap design across different model scales. Figure~\ref{fig:exp_async} reports the validation score under synchronous update and asynchronous updates with varying intervals ($I=1,3,5$).

For Qwen2.5-3B and 7B, we observe a clear advantage of asynchronous updates at small intervals. Specifically, Async-1 achieves the highest reward ($0.60$ and $0.42$, respectively), outperforming synchronous updates by a notable margin. Increasing the interval to 3 steps eliminates this gain, and at 5 steps performance further degrades below synchronous levels. This trend indicates that smaller models rely more critically on timely draft model adaptation to follow the evolving target distribution.

For Qwen2.5-14B, synchronous updates already achieve a relatively high reward ($0.60$), and Async-1 maintains this level without improvement. However, performance still degrades under longer intervals (Async-3: $0.50$, Async-5: $0.45$). This suggests that larger models are inherently more robust to stale draft updates, but still benefit from avoiding excessive delays.

Overall, these results validate our design principle: overlapping asynchronous updates with rollouts is most effective when updates are frequent (e.g., Async-1). This balance ensures the draft model remains aligned with target distribution while avoiding system stalls from synchronous training.


\section{Conclusion}

We present \SysName, the first system that adapts speculative decoding to reinforcement learning for large language models. By addressing key bottlenecks—diminishing speedups, drafter staleness, and policy degradation—\SysName integrates adaptive decoding, drafter evolution, and reward-weighted adaptation. Experiments on Qwen models show up to $4.5\times$ faster training with stable reward convergence, demonstrating the practicality of \SysName for scalable and efficient RL training.


\bibliography{references.bib}
\bibliographystyle{mlsys2025}

\appendix

\end{document}